\documentclass[10pt,twocolumn,letterpaper]{article}

\usepackage{iccv}
\usepackage{times}
\usepackage{epsfig}
\usepackage{graphicx}
\usepackage{amsmath}
\usepackage{amssymb}

\usepackage{booktabs}
\usepackage{bm}
\usepackage{multirow}
\usepackage{array}
\newcommand{\PreserveBackslash}[1]{\let\temp=\\#1\let\\=\temp}
\newcolumntype{C}[1]{>{\PreserveBackslash\centering}p{#1}}
\newcolumntype{R}[1]{>{\PreserveBackslash\raggedleft}p{#1}}
\newcolumntype{L}[1]{>{\PreserveBackslash\raggedright}p{#1}}

\newcommand{\sse}{*}
\newcommand{\ssv}{\sharp}

\usepackage[pagebackref=true,breaklinks=true,letterpaper=true,colorlinks,bookmarks=false]{hyperref}

\iccvfinalcopy 

\def\iccvPaperID{2227} 

\ificcvfinal\fi

\begin{document}

\title{ Boosting Entity-aware Image Captioning with Multi-modal Knowledge Graph }

\author{
   Wentian Zhao$^1$, Yao Hu$^2$, Heda Wang$^2$, Xinxiao Wu$^1$\thanks{Corresponding author: Xinxiao Wu}, and Jiebo Luo$^3$ \\
   $^1$ School of Computer Science, Beijing Institute of Technology, Beijing 100081, China\\
   $^2$ Alibaba Group\\
   $^3$ Department of Computer Science, University of Rochester, Rochester NY 14627, USA
}

\maketitle
\ificcvfinal\thispagestyle{empty}\fi

\begin{abstract}
Entity-aware image captioning aims to describe named entities and events related to the image by utilizing the background knowledge in the associated article.
This task remains challenging as it is difficult to learn the association between named entities and visual cues due to the long-tail distribution of named entities.
Furthermore, the complexity of the article brings difficulty in extracting fine-grained relationships between entities to generate informative event descriptions about the image.
To tackle these challenges, we propose a novel approach that constructs a multi-modal knowledge graph to associate the visual objects with named entities and capture the relationship between entities simultaneously with the help of external knowledge collected from the web.
Specifically, we build a text sub-graph by extracting  named entities and their relationships from the article, and build an image sub-graph by detecting the objects in the image.
To connect these two sub-graphs, we propose a cross-modal entity matching module  trained using a knowledge base that contains Wikipedia entries and the corresponding images.
Finally, the multi-modal knowledge graph is integrated into the captioning model via a graph attention mechanism.
Extensive experiments on both GoodNews and NYTimes800k datasets demonstrate the effectiveness of our method.

\end{abstract}

\section{Introduction}

\begin{figure}[t]
   \begin{center}
      \includegraphics[width=0.9\linewidth]{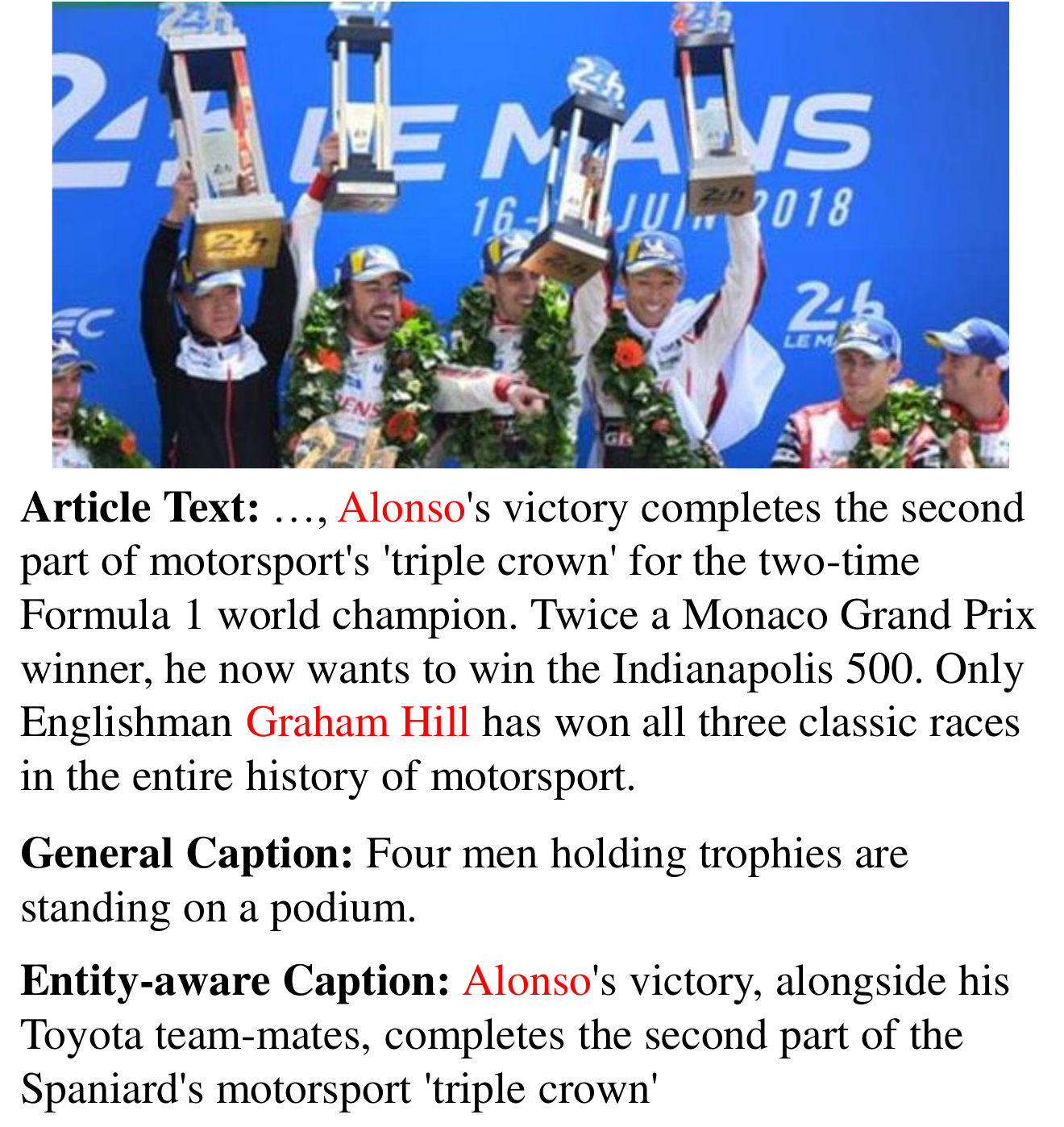}
   \end{center}
      \caption{An example of entity-aware image captioning. The named entities in type ``PERSON'' are marked in red.}
   \label{fig1}
\end{figure}

Different from the conventional image captioning \cite{vinyals2015show, karpathy2015deep, anderson2018bottom, huang2019attention, cornia2020meshed} that describes common objects and their relationships, entity-aware image captioning focuses on generating informative descriptions of named entities and specific events presented in the images by utilizing the background knowledge in the associated articles.
For instance, the image in Figure \ref{fig1} shows the scene where a famous racer is celebrating his victory.
A conventional captioning model may describe the general semantic in the image such as ``Four men are holding trophies on a podium.'', while the entity aware captioning model can leverage the relevant background knowledge to generate more expressive description such as ``Alonso is celebrating victory with his Toyota teammates''.
Entity-aware image captioning is closer to the human cognition process that integrates prior knowledge for understanding and interpreting scenes~\cite{wynn2020effects}, and has attracted increasing attention in the fields of computer vision and natural language processing \cite{ramisa2017breakingnews,lu2018entity,biten2019good,tran2020transform,hu2020icecap}.
However, this task still presents two key challenges.
First, since the distribution of named entities is heavily imbalanced and long-tailed, it is difficult to select the named entities relevant to the image from the articles.
Second, the news articles typically contain complex sentences where related entities might be far apart, making it non-trivial to exploit the fine-grained relationships between the named entities for generating descriptions about the events in the image.

\begin{figure*}[t]
   \begin{center}
   \includegraphics[width=1.0\linewidth]{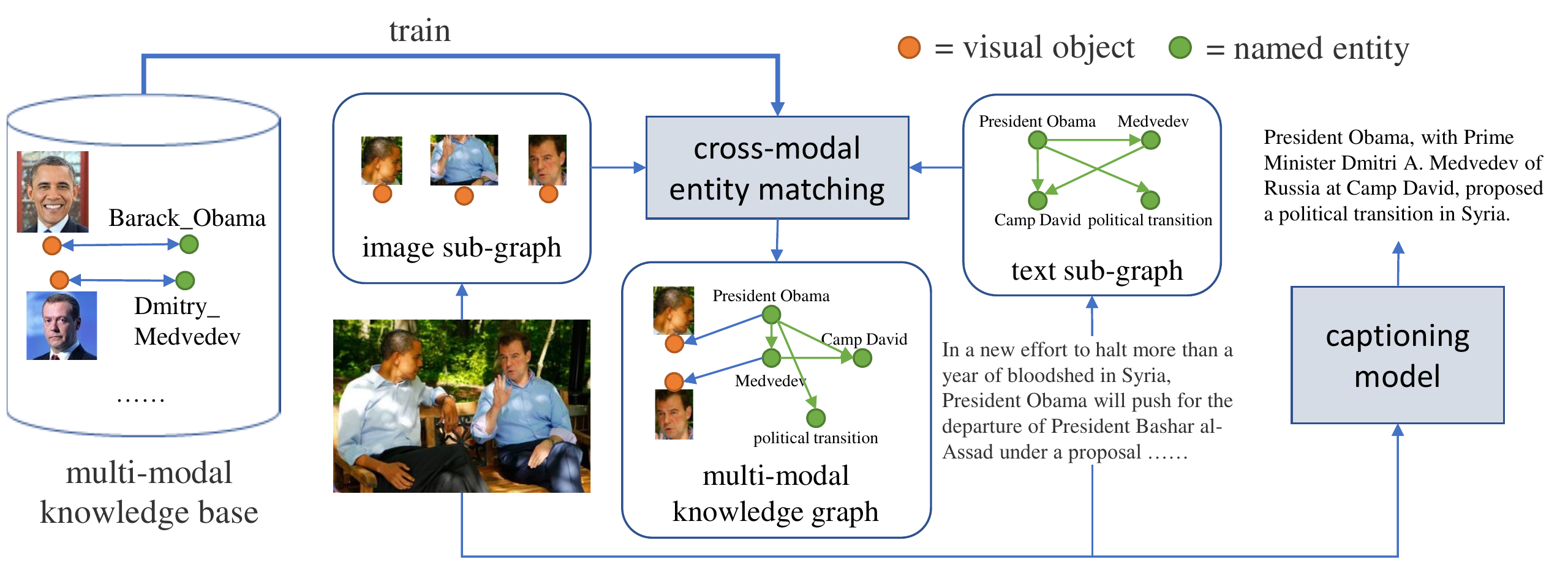}
   \end{center}
      \caption{The framework of our proposed method. The left part shows the external multi-modal knowledge base containing named entities and their corresponding images, which are used to train the cross-modal entity matching module. The middle part shows the generation process of the multi-modal knowledge graphs. An image sub-graph and a text sub-graph are extracted from the input image and the article text, respectively. The multi-modal entity matching module connects the related entities in the two sub-graphs to construct the multi-modal knowledge graph. The right part shows the captioning model, which encodes the image, the article and the multi-modal knowledge graph to generate an entity-aware caption. }
   \label{figure2}
\end{figure*}

There have been some attempts to select accurate named entities for the caption.
The template-based methods \cite{ramisa2017breakingnews, biten2019good} first generate template captions with placeholders, and then fill the placeholders using the named entities in the relevant sentences.
Several end-to-end methods have also been proposed, including using word-level contextual information to draw named entities~\cite{hu2020icecap} and applying byte-pair encoding to generate rare words in named entities~\cite{tran2020transform}.
The aforementioned named entity selection strategies heavily depend on the contextual information of the named entities in the article but neglect the association between named entities and visual cues in the image.

Recently there have been a few endeavours to utilize the background knowledge in the articles for concrete event description.
Many methods extract the knowledge by encoding the article text at article level \cite{ramisa2017breakingnews}, sentence level \cite{biten2019good} or word level \cite{tran2020transform}.
In \cite{hu2020icecap}, a more fine-grained attention mechanism is proposed to progressively concentrate on the text information from the sentence level to the word level.
Most of these methods employ sequence encoding models for captioning and lack ability to capture the entity relationships. This limits their applications to the difficult cases where the article presents more complex events with multiple named entities in different sentences. 

To overcome these limitations, we propose a multi-modal knowledge graph to explicitly model the association between visual objects and named entities and simultaneously capture the fine-grained relationships between named entities for entity-aware image captioning.
Since there are a wide range of named entities involved in the articles and the distribution of real-world named entities is often long-tailed, it is extremely difficult to learn the association between visual objects and named entities from the training data.
Therefore, we attempt to explore the external knowledge from the web which provides rich and comprehensive multi-modal information about named entities.
To be more specific, starting with collecting an external multi-modal knowledge base from Wikipedia, we then train a cross-modal entity matching module that connects a textual sub-graph extracted from the article and an image sub-graph extracted from the image, in order to construct the multi-modal knowledge graph. 
Finally, the multi-modal knowledge graph along with the input image and the news article is encoded via a graph attention mechanism into a captioning model to generate the entity-aware descriptions. 

We conduct extensive experiments on two widely-used news image captioning datasets, GoodNews \cite{biten2019good} and NYTimes800k \cite{tran2020transform}, to evaluate the effectiveness of the proposed method. In summary, our contributions are as follows:
\begin{itemize}
   \item We propose a novel entity-aware captioning method that constructs a multi-modal knowledge to choose accurate named entities and refine relevant events for generating informative descriptions.
   \item We design a novel cross-modal entity matching module that is effectively trained using a multi-modal knowledge base collected from Wikipedia to facilitate modeling the association between visual objects and named entities. 
   \item Experiments on two large-scale news image captioning datasets have verified the superiority of our method compared with the state-of-the-art methods.
\end{itemize}

\section{Related Work}

\subsection{Entity-aware Image Captioning}
Entity-aware image captioning is attracting increasing attention in recent years.
Some existing methods generate entity-aware captions by utilizing hashtags or named entities retrieved from the web \cite{lu2018entity, zhao2019informative}.
Several other methods extract the background knowledge from the associated news articles for generating the named entities \cite{ramisa2017breakingnews, lu2018entity, biten2019good, hu2020icecap, tran2020transform, yang2020image}, which are closer to our method.
Among these methods, the early works \cite{ramisa2017breakingnews, lu2018entity, biten2019good} generate entity-aware captions by two separate steps: a template caption is first generated with placeholders indicating the entity types, and then the placeholders are filled by selecting the named entities from the news article.
For instance, Biten \etal~\cite{biten2019good} first generate a template caption by attending to both the image and the sentences in the news article, and then draw the named entities from the sentences with the highest attention weights.

More recently, a few end-to-end methods \cite{hu2020icecap, tran2020transform, yang2020image} are proposed to generate entity-aware captions in one pass.
Tran \etal~\cite{tran2020transform} encode the news article at word level using a pre-trained language model, and handle the rare words in the named entities using byte-pair encoding \cite{sennrich2015neural}.
Hu \etal~\cite{hu2020icecap} first retrieve the sentences that are most relevant to the image, and then attend to the retrieved sentences.

Although remarkable progress has been achieved through leveraging textual information from the news article at the sentence or word level in the aforementioned methods, the explicit association between the named entities and visual objects is still under-explored, as well as the fine-grained relationship between named entities.
This paper focuses on constructing a multi-modal knowledge graph to associate the visual objects with named entities and capture the entity relationships, simultaneously, which is important for generating informative captions.

\subsection{Multi-modal Knowledge Graph}
Multi-modal knowledge graphs introduce information in multiple modalities, such as images, videos or text, to represent the entities and relations. 
Several recent studies have validated the effectiveness of multi-modal knowledge graphs in different fields.
For instance, some methods \cite{xie2016image, mousselly2018multimodal} incorporate images as additional features of the entities in the knowledge graph to learn better entity representations for knowledge graph completion and triple classification.
Kannan \etal~\cite{kannan2020multimodal} construct multi-modal knowledge base to excavate the facts in deep learning literatures.
A multi-modal knowledge graph based recommendation system is proposed in~\cite{sun2020multi}, where images and text entities are introduced to model user behaviour.

To the best of our knowledge, our method is the first to employ multi-modal knowledge graphs in entity-aware image captioning.
Compared with the methods mentioned above, constructing multi-modal knowledge graphs for the images and the associated news articles is more challenging since the relationship between the entities in the news article and the visual objects in the image is unknown.
Therefore, we leverage the external knowledge from the web to train a cross-modal entity matching module that establishes the connections between entities in different modalities.

\section{Our Method}

\subsection{Overview}
In this paper, we propose a multi-modal knowledge graph (MMKG) for entity-aware image captioning. It explicitly models the association between named entities and visual objects in the image and simultaneously captures the fine-grained relationships between named entities in the article. 
MMKG consists of a text sub-graph and an image sub-graph.
The text sub-graph models the interaction between the named entities in the article text, where the nodes represent the named entities and the directed edges represent the entity relationships.
The image sub-graph represents the visual objects detected in the image. To connect these two sub-graphs for generating the complete knowledge graph, we introduce a cross-modal entity matching module that measures the similarity between the named entities in the text sub-graph and the visual objects in the image sub-graph.
To facilitate the training of the cross-modal entity matching module, we collect an external multi-modal knowledge base from Wikipedia. Finally, the captioning model encodes the multi-modal knowledge graph as well as the image and article to generate the entity-aware captions. 
Figure \ref{figure2} illustrates the framework of our method.

\subsection{External Multi-modal Knowledge Base}
The external multi-modal knowledge base contains pairs of named entities and images, formulated as $D^M = \{({e}_i, {v}_i) | _i\}$, where ${e}_i$ and ${v}_i$ denote the named entity and the corresponding image, respectively.
The named entities in $D^M$ are collected from the news articles in the training splits of the GoodNews and NYTimes800k datasets. 
We perform named entity recognition on all the articles and keep the concrete entities, including persons, organizations, artifacts and facilities.
The abstract named entities, e.g. numbers and dates, are not included.
To reduce the ambiguity in the original text, we perform entity linking \cite{spitkovsky2012cross} to connect the recognized named entities in the news article to Wikipedia entries. 

Ideally, the image corresponding to a named entity should be representative, i.e. the most salient object in the images reflects the entity. 
For instance, the image that best reflects a person is the person's portrait, and the image corresponding to a building is the close-up of the building.
Note that the purpose of the images in each Wikipedia page is to demonstrate the named entity itself, and we believe that the images in $D^M$  collected from Wikipedia are with superior representativeness.
For each entity ${e}_i$, we use the first image in the corresponding Wikipedia page as ${v}_i$ for the sake of simplicity.
The multi-modal knowledge base contains 126,295 pairs of named entities and images in total.

\subsection{Cross-modal Entity Matching Module}
The cross-modal entity matching module measures the similarity between a named entity ${e}_i$ and an image ${v}_j$.
A pre-trained language model and a pre-trained CNN are used to encode the vector representations of ${e}_i$ and ${v}_j$, respectively, denoted as $\bm{u}_{{e}_i}$ and $\bm{u}_{{v}_j}$.
The cross-modal entity matching module is trained to map the vectors  $\bm{u}_{{e}_i}$ and $\bm{u}_{{v}_j}$ into a common embedding space, where the similarity between positive entity-image pairs is larger than any negative pairs by a margin $\delta$.

We use the pairs of named entities and images in the multi-modal knowledge base to train the cross-modal entity matching module.
The loss function used to train the cross-modal entity matching module is formulated as
\begin{equation}
   \begin{aligned}
      L_r = & \max_{{e}'}(\delta + {\rm sim}({e}', {v}) - {\rm sim}({e}, {v}))_+ \\+ & \max_{{v}'}(\delta + {\rm sim}({e}, {v}') - {\rm sim}({e}, {v}))_+,
   \end{aligned}
\end{equation}
where the pairs $({e}', {v})$ and $({e}, {v}')$ denote negative samples, ${\rm sim}({e}, {v})$ denotes the similarity between ${e}$ and ${v}$, and $(x)_+ = {\rm max}(x, 0)$.

\subsection{Multi-modal Knowledge Graph}
\begin{figure}[t]
   \begin{center}
      \includegraphics[width=1.0\linewidth,height=7.7cm]{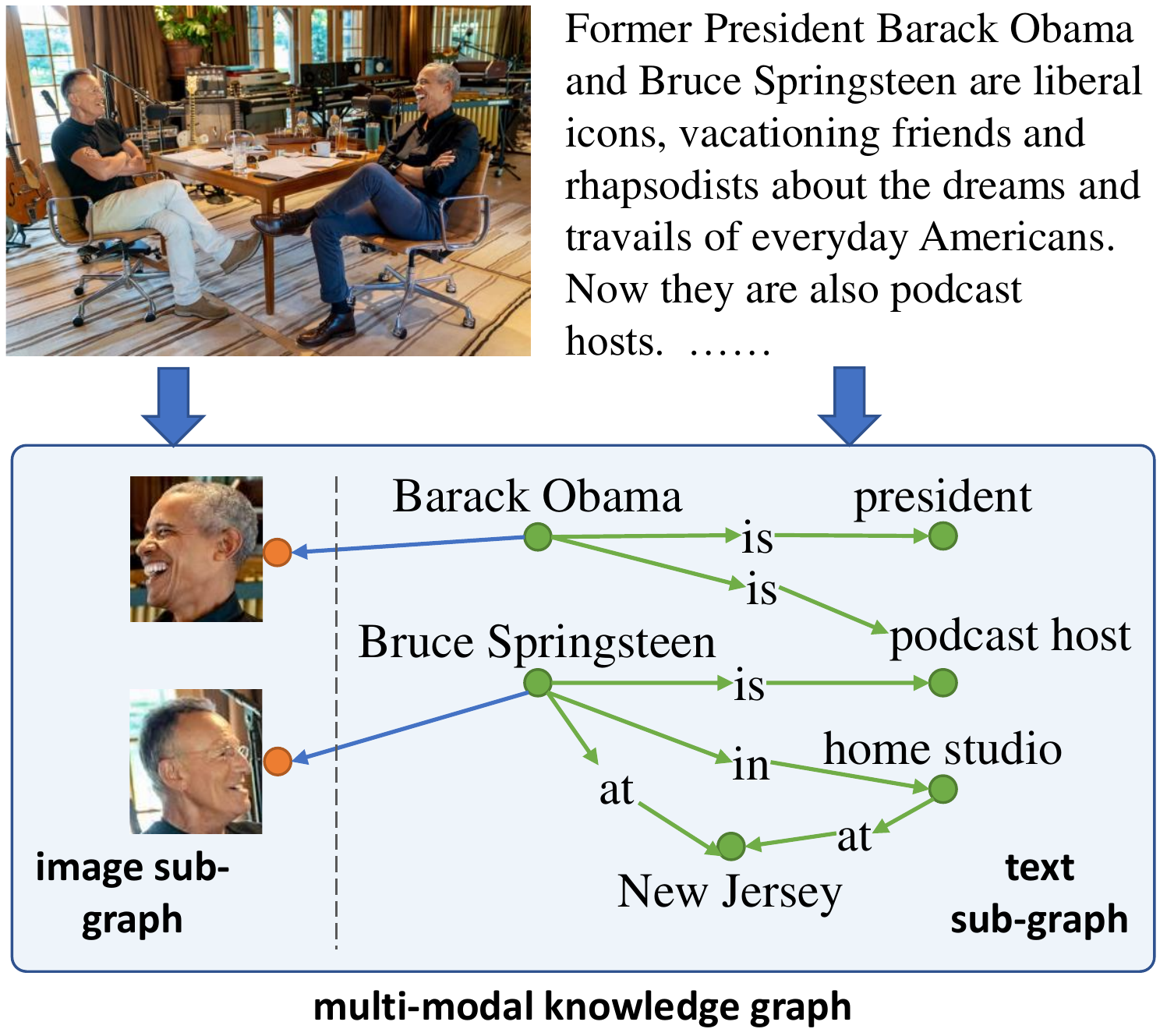}
   \end{center}
      \caption{An example of the constructed multi-modal knowledge graph, which consists of an image sub-graph (the left part of the box) and a text sub-graph (the right part of the box).}
   \label{fig4}
\end{figure}

Compared with conventional knowledge graphs that only contains entities extracted from text, multi-modal knowledge graphs introduce additional modality of entities and relationships, such as images.
Formally, a multi-modal knowledge graph is denoted as $G^M =\langle V, E \rangle$, where $V$ denotes the entity set and $E$ denotes the edge set.
The entity set $V$ contains both the entities from the article and the visual objects from the image.
We construct the multi-modal knowledge graph by first building a text sub-graph and an image sub-graph, and then connecting these two sub-graphs using the cross-modal entity matching module.
An example multi-modal knowledge graph is shown in Figure \ref{fig4}.

To build the text sub-graph, denoted as $G^T=\langle V^T, E^T \rangle$, where $V^T$ and $E^T$ denote the entities in the news article and the edges that connects them, we perform information extraction and coreference resolution by utilizing the Stanford CoreNLP toolkit \cite{manning2014stanford}.
Specifically, for an article $T$, the output of information extraction is a set of triples $R^T = \{ \langle e^h_i, e^r_i, e^t_i \rangle | _i \}$, where $e^h_i$, $e^r_i$ and $e^t_i$ denote the head entity, the relation and the tail entity, respectively.
For each triple $\langle e^h_i, e^r_i, e^t_i \rangle$, we create two directed edges $e^h_i \rightarrow  e^r_i$ and $e^r_i \rightarrow e^t_i$.
To simplify the graph structure, for the same entity that appears more than once in the article, we only keep the entity that appears first.
For instance, for two triples $\langle e^h_i, e^r_i, e^t_i \rangle$ and $\langle e^h_j, e^r_j, e^t_j \rangle$ where $e^h_i$ and $e^h_j$ refer to the same entity and $e^h_i$ appears first in the article text, we remove $e^h_j$ from the knowledge graph and create an edge $e^h_i \rightarrow e^r_j$.

To construct the image sub-graph, denoted as $G^I=\langle V^I, E^I \rangle$, we use the YOLOv3 object detector \cite{redmon2018yolov3} and the MTCNN \cite{zhang2016joint} network to detect the objects and faces in the image, and the detected objects and faces are denoted by $V^o = \{v_i^o | _i\}$ and $V^f = \{v_i^f | _i\}$, respectively.
The entity set $V^I$ of the image sub-graph contains both the set of objects and the set of faces, i.e. $V^I = V^o \cup V^f$.
For each object or face $v_i^{\ssv}$, we obtain its vector representation $\bm{u}_{v_i^\ssv} \in \mathbb{R}^{2048}$ using the pre-trained ResNet-152 model \cite{he2016deep}, where $\ssv \in \{o, f\}$.

The similarities between the entities in the image sub-graph and the text sub-graph are measured by the cross-modal entity matching module.
For all the entity pairs $e_i^{\sse}, v_j^{\ssv}$ that satisfies ${\rm sim}(e_i^{\sse}, v_j^{\ssv}) > 0.4$, we create a directed edge $e_i^{\sse} \rightarrow v_j^{\ssv}$ that connects them, where $\sse \in \{h, r, t\}$.

\subsection{Entity-aware Captioning Model}
The generated multi-modal knowledge graph $G^M$, the input image $I$ and the associated news article $T$ are encoded by the encoder of the entity-aware captioning model. 
The image $I$ is encoded using the pre-trained ResNet-152, and the output before the last pooling layer is taken and flattened into a matrix $\bm{X}^I = \{\bm{x}^I_j | _j \}$, $\bm{x}^I_j \in \mathbb{R}^{2048}$.

The news article $T$ is encoded by the pre-trained RoBERTa~\cite{liu2019roberta} into a sequence of subword units $\{w_1, w_2, ..., w_{L_T}\}$, where $L_T$ denotes the length of the subword unit sequence.
The output of the last layer is used as the representation, denoted as $\bm{X}^T = \{\bm{x}^T_j | _j\}$, $\bm{x}^T_j \in \mathbb{R}^{1024}$.

To obtain rich representations of the entities in the multi-modal knowledge graph, we encode the nodes in the multi-modal knowledge graph using a two-layer graph attention network (GAT) \cite{velivckovic2017graph}.
The node representations in the multi-modal knowledge graph, denoted as $\bm{U}^G = \bm{U}^T \cup \bm{U}^I$, is used as the input of the GAT.
$\bm{U}^T=\{\bm{u}_{e_i^\sse} | e_i^\sse \in V^T\}$ and $\bm{U}^G=\{\bm{u}_{v_j^\ssv} | v_j^\ssv \in V^I\}$ denote the vector representations of the nodes in the text sub-graph $G^T$ and the image sub-graph $G^I$, respectively, where $\bm{u}_{e_i^\sse}$ and $\bm{u}_{v_j^\sse}$ denote the initial vector representations of named entity $e_i^\sse$ and visual object $v_j^\ssv$, respectively.
The output of the last GAT layer is used as the input to the decoder, denoted as $\bm{X}^G = \{\bm{x}_{e_i^\sse} | e_i^\sse \in V^T\} \cup \{\bm{x}_{v_j^\ssv} | v_j^\ssv \in V^I\}$ where $\bm{x}_{e_i^\sse}$ and $\bm{x}_{v_j^\ssv}$ denote the vector representations of named entity $e_i^\sse$ and visual object $v_j^\ssv$ that are encoded by the graph attention network, respectively.

\begin{table*}[t]
   \begin{center} 
     
     \begin{tabular}{c | l | R{0.08\textwidth}R{0.08\textwidth}R{0.08\textwidth}R{0.08\textwidth}R{0.08\textwidth}}
      \toprule
      \multicolumn{1}{c|}{Dataset} & Method & \multicolumn{1}{r}{Bleu-4} & \multicolumn{1}{r}{METEOR} & \multicolumn{1}{r}{ROUGE} & \multicolumn{1}{r}{CIDEr} & \multicolumn{1}{r}{entity F1} \\
      \midrule
      \multirow{8}[2]{*}{GoodNews} & SAT+CtxIns \cite{xu2015show} & 0.38  & 3.56  & 10.50  & 12.09  & 6.09  \\
            & TopDown + CtxIns \cite{anderson2018bottom} & 0.70  & 3.81  & 11.09  & 13.38  & 6.87  \\
            & Ramisa et al. + CtxIns \cite{ramisa2017breakingnews} & 0.89  & 4.45  & 12.09  & 15.35  & 7.54  \\
            & Biten et al. + CtxIns \cite{biten2019good} & 0.69  & 4.14  & 11.70  & 13.77  & 7.19  \\
            & Biten et al. + AttIns & 0.76  & 4.02  & 11.44  & 13.69  & 7.97  \\
            & ICECAP \cite{hu2020icecap} & 1.96  & 6.01  & 15.70  & 26.08  & 12.03  \\
            & Transform and Tell \cite{tran2020transform} & 6.05  &     - & 21.40  & 53.80  & 20.30  \\
            & Ours  & \textbf{6.14} & \textbf{6.32} & \textbf{21.46} & \textbf{54.02} & \textbf{20.36} \\
      \midrule
      \multirow{2}[2]{*}{NYTimes800k} & Transform and Tell \cite{tran2020transform} & 6.30  &    -  & \textbf{21.70} & 54.40  & 23.33 \\
            & Ours  & \textbf{6.32} & \textbf{6.25} & 21.62  & \textbf{54.47} & \textbf{23.44}  \\
      \bottomrule
      \end{tabular}%
   \end{center}
   \caption{Evaluation Results on the GoodNews dataset and the NYTimes800k dataset. The best results are marked in Bold.}
   \label{table1}%
\end{table*}%

The decoder of the captioning model generates the tokens in the entity-aware captions sequentially and consists of $N$ identical Transformer layers.
The initial input to the decoder is denoted as $\bm{X}_0 = [\bm{X}^T; \bm{X}^I; \bm{X}^G]$, where the operator $[ ; ]$ denotes matrix concatenation.
At the $t$-th time step, the decoder predicts the probability of the current token $\bm{p}_t \in \mathbb{R}^D$ using the initial input and the embeddings of the previously generated subword units $\bm{M}_{t-1} = \{\bm{m}_0, \bm{m}_1, ..., \bm{m}_{t-1}\}$, where $\bm{m}_i$ denotes the embedding of the $i$-th subword unit and $D$ denotes the vocabulary size.

During training of the captioning model, we fix the parameters of the pre-trained RoBERTa and the pre-trained CNN in the encoder. 
The parameters of  GAT  and the decoder are optimized using the following cross-entropy loss function:
\begin{equation}
   \begin{aligned}
      L_p = - \sum_{t=1}^{|Y|} \log p(w_t|w_1, w_2, ..., w_{t-1}),
   \end{aligned}
\end{equation}
where $|Y|$ denotes the length of the ground-truth caption, and $w_i$ denotes the $i$-th token in the caption $Y$.

\begin{table*}[t]   
   \begin{center}
   \begin{tabular}{c|l| R{0.08\textwidth}R{0.08\textwidth}R{0.08\textwidth}R{0.08\textwidth}R{0.08\textwidth} }
      \toprule
      \multicolumn{1}{c|}{Dataset} & Method & \multicolumn{1}{r}{Bleu-4} & \multicolumn{1}{r}{METEOR} & \multicolumn{1}{r}{ROUGE} & \multicolumn{1}{r}{CIDEr} & \multicolumn{1}{r}{entity F1} \\
      \midrule
      \multirow{4}[2]{*}{GoodNews} & w/o graph & 5.32  & 6.03  & 19.30  & 47.63  & 18.23  \\
            & image sub-graph & 5.41  & 5.97  & 19.54  & 47.52  & 18.87  \\
            & text sub-graph & 5.93  & 6.15  & 20.13  & 48.56  & 19.23  \\
            & Ours  & 6.14  & 6.32  & 21.46  & 54.02  & 20.36  \\
      \midrule
      \multirow{4}[2]{*}{NYTimes800k} & w/o graph & 5.26  & 5.87  & 19.71  & 48.49  & 19.72  \\
            & image sub-graph & 5.32  & 6.03  & 19.81  & 48.53  & 20.13  \\
            & text sub-graph & 6.03  & 6.41  & 21.30  & 52.76  & 22.03  \\
            & Ours  & 6.32  & 6.25  & 21.62  & 54.47  & 23.44  \\
      \bottomrule
   \end{tabular}%
   \end{center}
   \caption{The results of ablation studies on the GoodNews dataset and the NYTimes800k dataset. }
   \label{table2}%
 \end{table*}%

\section{Experiments}
\subsection{Datasets}
Our experiments are conducted on two news image captioning datasets, GoodNews \cite{biten2019good} and NYTimes800k \cite{tran2020transform}.
The images, captions and news articles in these datasets are collected from New York Times, and each image is annotated with one ground-truth caption.
In the GoodNews dataset, since the URLs to some images are not accessible, we employ the dataset split in \cite{tran2020transform}, where the train, validation and test splits contain 445,259 images, 19,448 images and 24,461 images, respectively.
The average lengths of the news articles and the ground-truth captions are 451 words and 18 words, respectively.
The NYTimes800k dataset contains 763,217 images, 7,777 images and 21,977 images for training, validation and testing, respectively.
Compared to the GoodNews dataset, NYTimes800k is more complex since the average article length and the average caption length of NYTimes800k are 974 words and 18 words, respectively. 

We use standard image captioning metrics, including Bleu-4 \cite{papineni2002bleu}, ROUGE-L \cite{lin2004rouge} and CIDEr \cite{vedantam2015cider}, to evaluate the similarity of the generated captions to the ground-truth captions.
Since the goal of the model is to generate entity-aware captions, we also evaluate the named entities in the genrated sentences.
Specifically, the Spacy toolkit \cite{spacy} is used to recognize the named entities in both ground-truth sentences and generated sentences.
We use the exact string matching to compare the generated and ground-truth named entities, and report the F1 score of the generated named entities.

\subsection{Implementation Details}

To detect the objects in the image $I$, we use the YOLOv3 detector \cite{redmon2018yolov3} and filter out the object bounding boxes with a confidence score of less than 0.3, while a maximum of 64 objects are kept for each image.
To detect the faces, we use the MTCNN \cite{zhang2016joint} network.
Following the practice in \cite{tran2020transform}, we keep no more than 4 face bounding boxes with the highest confidence scores for each image.
To extract the features of the input images as well as the detected objects and faces, we use the pre-trained ResNet-152 \cite{he2016deep} model.

The hidden dimension of the decoder in the captioning model is set to 1024.
The sub-word vocabulary of the decoder is identical to the sub-word vocabulary of the pre-trained RoBERTa \cite{liu2019roberta} model, which contains about 50,000 sub-word units.
We set the maximum length of the article text $T$ and the caption $Y$ to 512 tokens and 50 tokens, respectively.
We optimize the parameters of the model with the Adam optimizer \cite{kingma2014adam}.
We apply $L_2$ regularization to the parameters, and we use a weight decay of $10^{-5}$.
The initial learning rate is set to $10^{-7}$. 
We increase the learning rate to $10^{-4}$ in the first 4,000 warmup steps, and decrease the learning rate linearly in the following training steps.
We clip the gradients by the norm of $0.1$, and the batch size is set to 16.

\subsection{Comparison with State-of-the-Art Methods}
We compare our method with several types of entity-aware captioning methods, i.e. conventional image captioning methods \cite{xu2015show, anderson2018bottom}, and recent methods with associated articles including template-based methods \cite{ramisa2017breakingnews, biten2019good} and end-to-end methods \cite{biten2019good, tran2020transform, hu2020icecap}.
To evaluate the conventional models (SAT \cite{xu2015show} and TopDown \cite{anderson2018bottom}), the original image captioning model only takes the image as input and generate a template caption, and the placeholders are filled by comparing the contextual information of the placeholders and the named entities in the news article, denoted as ``CtxIns''.

Different form the conventional methods, the template-based methods (Ramisa \etal~\cite{ramisa2017breakingnews} and Biten \etal~\cite{biten2019good}) generate the template caption using both the image and the news article.
Ramisa \etal~\cite{ramisa2017breakingnews} encode the article into a single vector.
Biten \etal~\cite{biten2019good} attend to both the image regions and the sentences in the article, where two named entity selection strategies are employed, i.e. using sentence-level attention scores (denoted as ``AttIns'') and using contextual information (denoted as ``CtxIns'').

In contrast to the template-based methods that generate the captions in a two-stage manner, end-to-end methods \cite{tran2020transform, hu2020icecap} directly generate entity-aware captions in one pass.
Transform and Tell \cite{tran2020transform} is a Transformer-based \cite{vaswani2017attention} captioning model that uses byte-pair encoding to handle rare words and named entities.
ICECAP \cite{hu2020icecap} progressively attends to the sentences and the words in the article, and uses the contextual information to select the named entities.

Table \ref{table1} shows the comparison results between the aforementioned state-of-the-art methods and our method.
We observe that our method achieves the best results in terms of most evaluation metrics on both datasets, especially Bleu-4 and CIDEr, which validates the superority of the proposed multi-modal knowledge graph on entity-aware image captioning.
The entity F1 score of our method outperforms all existing methods on both datasets, which indicate that by modeling the association 
between the named entities and the visual cues in the image, our method selects named entities from the news articles more accurately.

\begin{table*}[t]
   \begin{center}
   \begin{tabular}{l|R{0.07\textwidth}R{0.07\textwidth}R{0.07\textwidth}R{0.07\textwidth}|R{0.07\textwidth}R{0.07\textwidth}R{0.07\textwidth}R{0.07\textwidth}}
      \toprule
      \multicolumn{1}{l|}{\multirow{2}[2]{*}{Method}} & \multicolumn{4}{c|}{GoodNews} & \multicolumn{4}{c}{NYTimes800k} \\
            & \multicolumn{1}{r}{Bleu-4} & \multicolumn{1}{r}{METEOR} & \multicolumn{1}{r}{ROUGE} & \multicolumn{1}{r|}{CIDEr} & \multicolumn{1}{r}{Bleu-4} & \multicolumn{1}{r}{METEOR} & \multicolumn{1}{r}{ROUGE} & \multicolumn{1}{r}{CIDEr} \\
      \midrule
      w/o graph & 5.27  & 5.59  & 17.32  & 40.45  & 5.37  & 5.65  & 17.23  & 42.83  \\
      image sub-graph & 5.29  & 5.53  & 17.93  & 41.26  & 5.41  & 5.34  & 18.10  & 42.82  \\
      text sub-graph & 5.72  & 5.78  & 18.21  & 43.37  & 5.74  & 5.96  & 18.21  & 46.59  \\
      ours  & 5.89  & 5.98  & 19.19  & 45.22  & 5.97  & 6.03  & 19.20  & 47.27  \\
      \bottomrule
      \end{tabular}%
   \end{center}
   \caption{Evaluation results of event description on the GoodNews dataset and the NYTimes800k dataset. To eliminate the influence of the named entities, they are replaced by the corresponding class labels.}
   \label{table3}%
\end{table*}%

\subsection{Ablation Studies}
To verify the effect of each component in our method, we conduct ablation studies on the GoodNews dataset.
We evaluate the following variants of our model:
\begin{itemize}
   \item \textbf{w/o graph}: To evaluate the effect of the multi-modal knowledge graph, the entire multi-modal knowledge graph is removed, and the input to the captioning model only includes the region features of the image $\bm{X}^I$ and the token-level features of the news article $\bm{X}^T$. 
   \item \textbf{image sub-graph}: To evaluate the contribution of the text sub-graph $G^T$, the text sub-graph is removed and the input to the decoder includes $\bm{X}^T$, $\bm{X}^I$ and the features of the objects and faces in the image.
   \item \textbf{text sub-graph}: To evaluate the contribution of the image sub-graph, we remove the objects and faces in $G^I$. The input to the decoder includes $\bm{X}^T$, $\bm{X}^I$ and the representations of the nodes in the text sub-graph $G^T$.
\end{itemize}

The results are shown in Table \ref{table2}.
From these results, we can make the following observations:
First, the performance of ``w/o graph'' significantly degrades on all the metrics, which indicate that the multi-modal knowledge graph is beneficial to describing the events in the image as well as selecting named entities.
Second, when using only the image sub-graph or the text sub-graph, the values of most metrics slightly increase, demonstrating that either sub-graphs contribute to improving the performance.
Third, compared to the image sub-graph, using the text sub-graph achieves better results, indicating that the background knowledge in the news article is of greater importance than the objects and faces in the image.
Finally, our full model performs best when the text sub-graph and the image sub-graph are both encoded, validating the effectiveness of learning the association between the named entities and the visual objects.

\begin{figure*}[t]
   \begin{center}
      \includegraphics[width=0.97\linewidth]{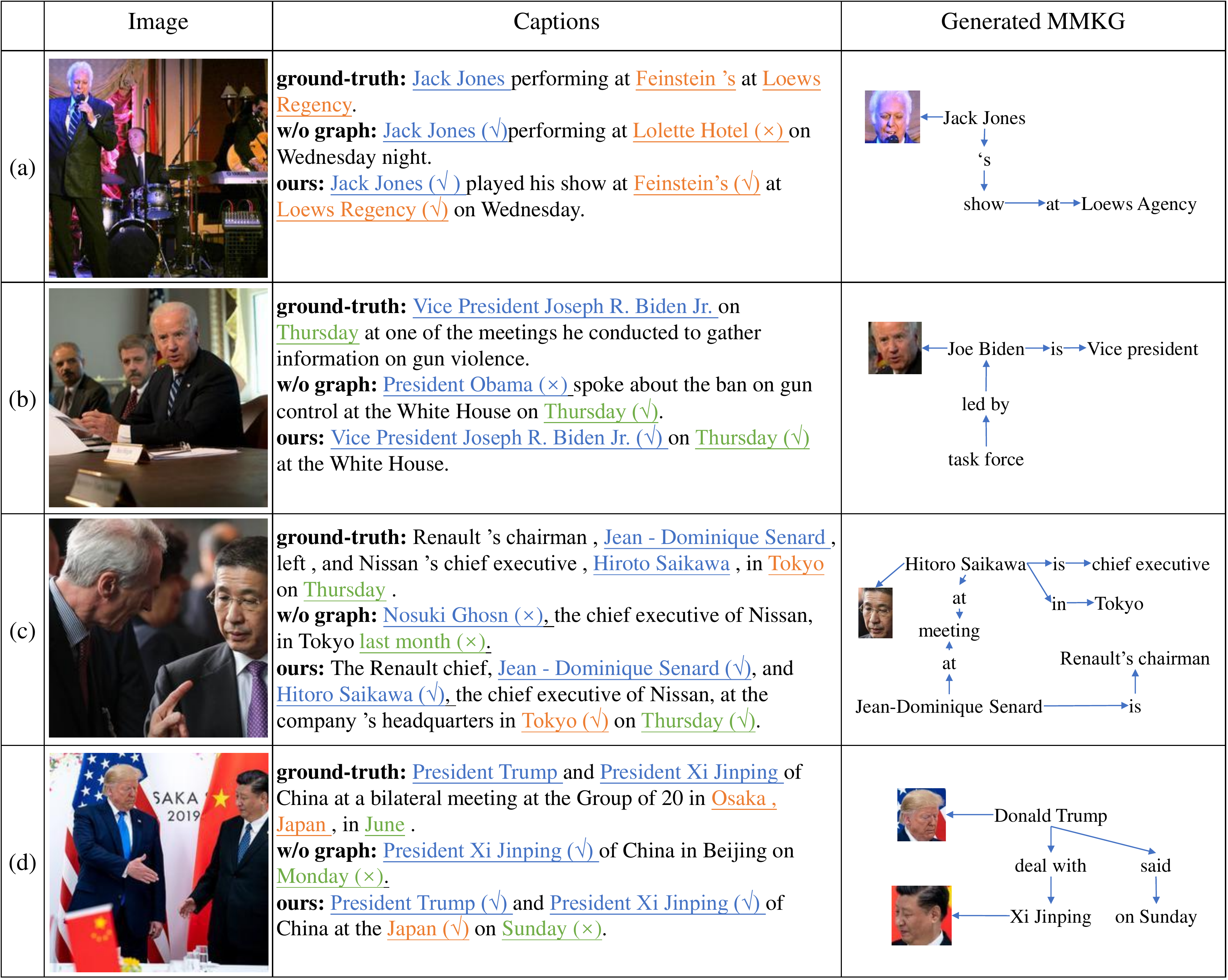}
   \end{center}
      \caption{Qualitative results on the GoodNews dataset ((a), (b)) and the NYTimes800k dataset ((c), (d)). ``ground-truth'', ``w/o graph'' and ``ours'' denote the ground-truth caption, the caption generated without using the multi-modal knowledge graph, and the caption generated by our method, respectively. The named entities in the captions are colored and underlined. Due to space limitation, the rightmost column only shows part of the constructed multi-modal knowledge graphs.}
   \label{fig3}
\end{figure*} 

\subsection{Evaluating Event Description Performance}
In addition to selecting named entities correctly, an entity aware captioning model is also required to describe the events depicted in the images properly.
To evaluate the performance of event description, the influence of named entities should be eliminated.
We perform named entity recognition using the SpaCy toolkit on both the ground-truth captions and generated captions, and replace the named entities with corresponding class labels.
The standard image captioning evaluation metrics, including Bleu-4, ROUGE-L and CIDEr, are reported.

The event description evaluation results on GoodNews and NYTimes800k are shown in Table \ref{table3}.
From these results, we observe that both the image sub-graph and the text sub-graph contribute to event description.
The model using the text sub-graph performs better than the model using image sub-graph, which indicates that the named entity relationships in the text sub-graph play a more important role in describing the events in the image.

\subsection{Qualitative Results}
We show some exemplars of generated entity-aware captions in Figure \ref{fig3}.
As illustrated in the figure, our model selects multiple types of named entities more accurately, including concrete entities including  persons, as well as abstract entities, such as places and times.
For example, in Figure \ref{fig3}(b), our method correctly describes the person name in the image, while the model without multi-modal knowledge graph uses another person name in the article.
The cases where multiple named entities are related to the image is also well handled by our model, thanks to the modeling of entity relationships using the multi-modal knowledge graphs.
For instance, in Figure \ref{fig3}(d) where the image shows multiple persons, our model successfully identifies all the persons.
It is also interesting to observe that even if the visual objects are not clear, our method still works well, which further validates the advantage of the entity relationship captured by the multi-modal graph using the external knowledge.
Taking Figure \ref{fig3}(c) for example, though the Renault's chairman Jean-Dominique Senard (the person on the left) is shadowed and is difficult to recognize, our model correctly describes both person names using the relationships involving him and the Nissan's executive, Hitoro Saikawa (the person on the right).

\section{Conclusion}
We present a novel entity-aware image captioning method that constructs a multi-modal knowledge graph by exploring external knowledge from the web. 
Our method can simultaneously associate visual cues with named entities and capture the fine-grained relationships between named entities, thus succeeding in extracting accurate entities and refining concrete events.
The proposed method first constructs a text sub-graph that consists of the named entities and their relationships, and an image sub-graph containing the visual objects in the image, and then connects the similar named entities and visual objects using a cross-modal entity matching module.
Extensive experiments on two large-scale entity-aware image captioning datasets, GoodNews and NYTimes800k, demonstrate the effectiveness of the proposed method.
In the future, we are going to investigate more advanced caption decoders and linguistic features to further improve the captioning performance.

{\small
\bibliographystyle{ieee_fullname}
\bibliography{egbib}
}

\end{document}